\documentclass[10pt,twocolumn,letterpaper]{article}

\usepackage{iccv}
\usepackage{times}
\usepackage{epsfig}
\usepackage{graphicx}
\usepackage{amsmath}
\usepackage{amssymb}
\usepackage{booktabs}
\usepackage[accsupp]{axessibility}  

\usepackage[pagebackref=true,breaklinks=true,letterpaper=true,colorlinks,bookmarks=false]{hyperref}

\iccvfinalcopy 

\def\iccvPaperID{****} 

\ificcvfinal\fi

\begin{document}

\title{What Does TERRA-REF's High Resolution, Multi Sensor Plant Sensing Public Domain Data Offer the Computer Vision Community?}

\author{David LeBauer\\
University of Arizona\\
{\tt\small dlebauer@arizona.edu}
\and
Max Burnette\\
University of Illinois \\ 
{\tt\small burnette@illinois.edu}\\
\and 
Noah Fahlgren \\ 
Donald Danforth Plant Science Center \\ 
{\tt\small nfahlgren@danforthcenter.org}
\and 
Rob Kooper \\ 
University of Illinois \\ 
{\tt\small kooper@illinois.edu}
\and Kenton McHenry \\
University of Illinois \\  
{\tt\small mchenry@illinois.edu}
\and Abby Stylianou \\ 
St. Louis University \\ 
{\tt\small abby.stylianou@slu.edu}
}

\maketitle
\ificcvfinal\thispagestyle{empty}\fi

\begin{abstract}
A core objective of the TERRA-REF project was to generate an open-access reference dataset for the evaluation of sensing technologies to study plants under field conditions. 
The TERRA-REF program deployed a suite of high-resolution, cutting edge technology sensors on a gantry system with the aim of scanning 1 hectare (10\textsuperscript{4}m) at around 1 mm\textsuperscript{2} spatial resolution multiple times per week.
The system contains co-located sensors including a stereo-pair RGB camera, a thermal imager, a laser scanner to capture 3D structure, and two hyperspectral cameras covering wavelengths of 300-2500nm.
This sensor data is provided alongside over sixty types of traditional plant phenotype measurements that can be used to train new machine learning models.
Associated weather and environmental measurements, information about agronomic management and experimental design, and the genomic sequences of hundreds of plant varieties have been collected and are available alongside the sensor and plant phenotype data.

Over the course of four years and ten growing seasons, the TERRA-REF system generated over 1 PB of sensor data and almost 45 million files. 
The subset that has been released to the public domain accounts for two seasons and about half of the total data volume.
This provides an unprecedented opportunity for investigations far beyond the core biological scope of the project.

The focus of this paper is to provide the Computer Vision and Machine Learning communities an overview of the available data and some potential applications of this one of a kind data.

\end{abstract}

\section{Introduction}

\begin{figure}[hbt]
\includegraphics[width=0.45\textwidth]{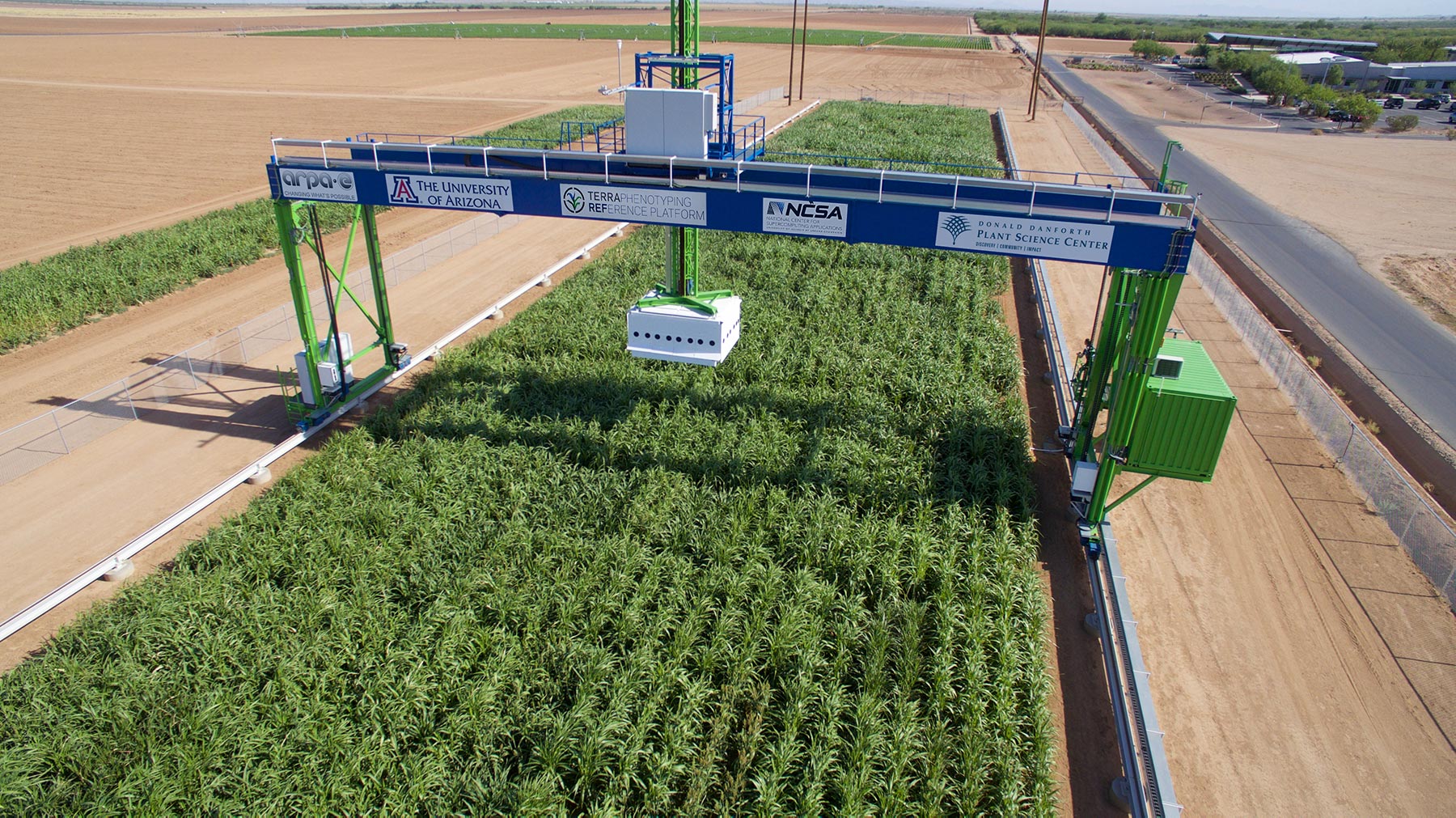}
\caption{TERRA-REF field scanner at the University of Arizona's Maricopa Agricultural Center.}
\label{fig:field_scanner}
\end{figure}

In 2015, the Advanced Research Projects Agency for Energy (ARPA-E) funded the TERRA-REF Phenotyping Platform (Figure ~\ref{fig:field_scanner}). 
The scientific aim was to transform plant breeding by providing a reference dataset generated by deploying a suite of co-located high-resolution sensors under field conditions.
The goal of these sensors was to use proximate sensing from approximately 2m above the plant canopy to quantify plant characteristics.

The study has evaluated diverse populations of sorghum, wheat, and lettuce over the course of four years and ten cropping cycles.
Future releases of additional data will be informed by user interests.

The TERRA-REF reference dataset can be used to characterize phenotype-to-genotype associations, on a genomic scale, that will enable knowledge-driven breeding and the development of higher-yielding cultivars of sorghum and wheat.
The data is also being used to develop new algorithms for machine learning, image analysis, genomics, and optical sensor engineering.
Beyond applications in plant breeding, the resulting dataset provides opportunities for the study and integration of diverse remote sensing modalities.

\subsection{Types of Data}

\begin{figure}[hbt]
\centering
\includegraphics[width=0.45\textwidth]{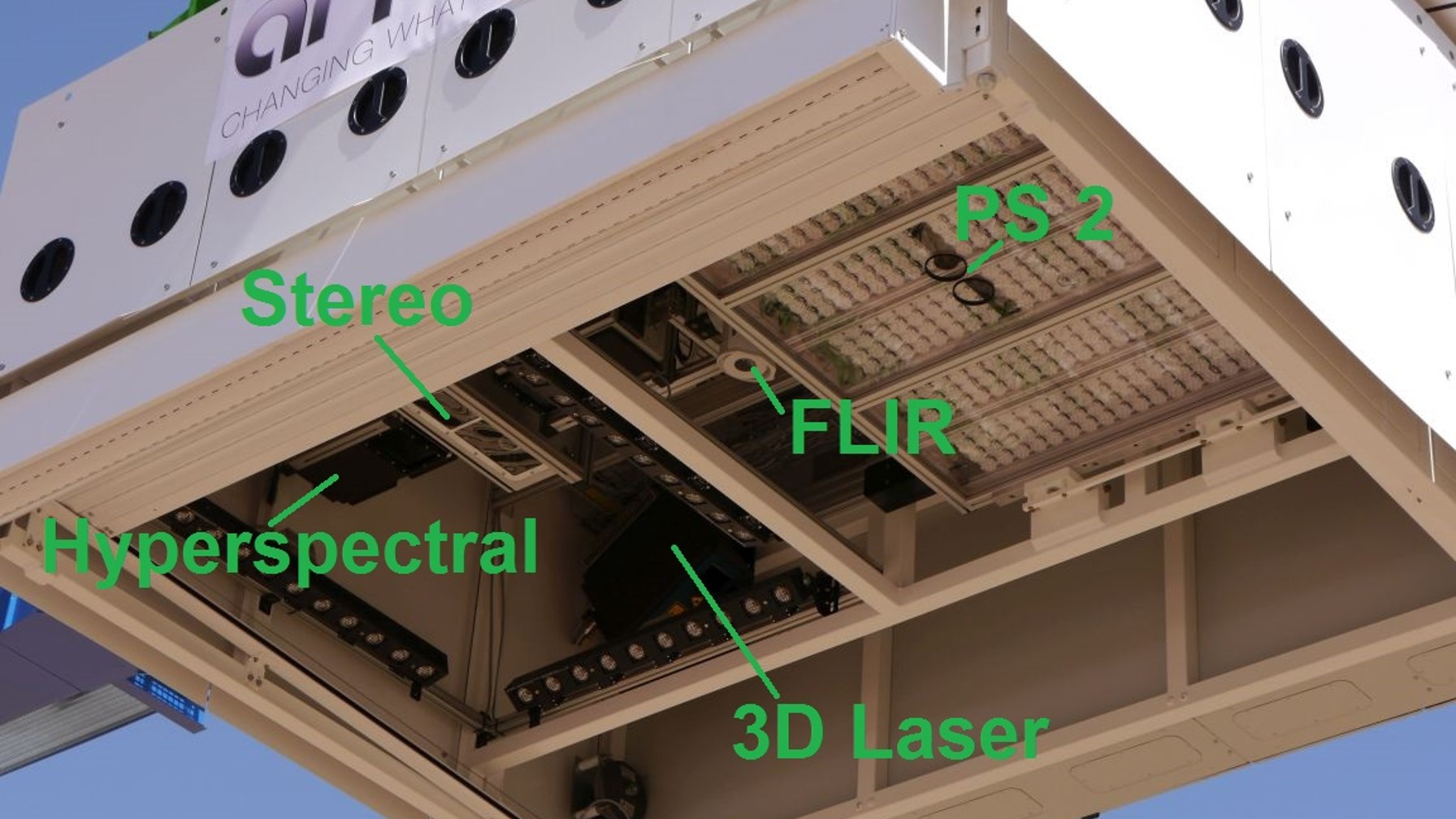}
\caption{TERRA-REF field scanner sensor suite.}
\label{fig:sensors}
\end{figure}

\begin{figure*}
\centering
\includegraphics[width=0.8\textwidth]{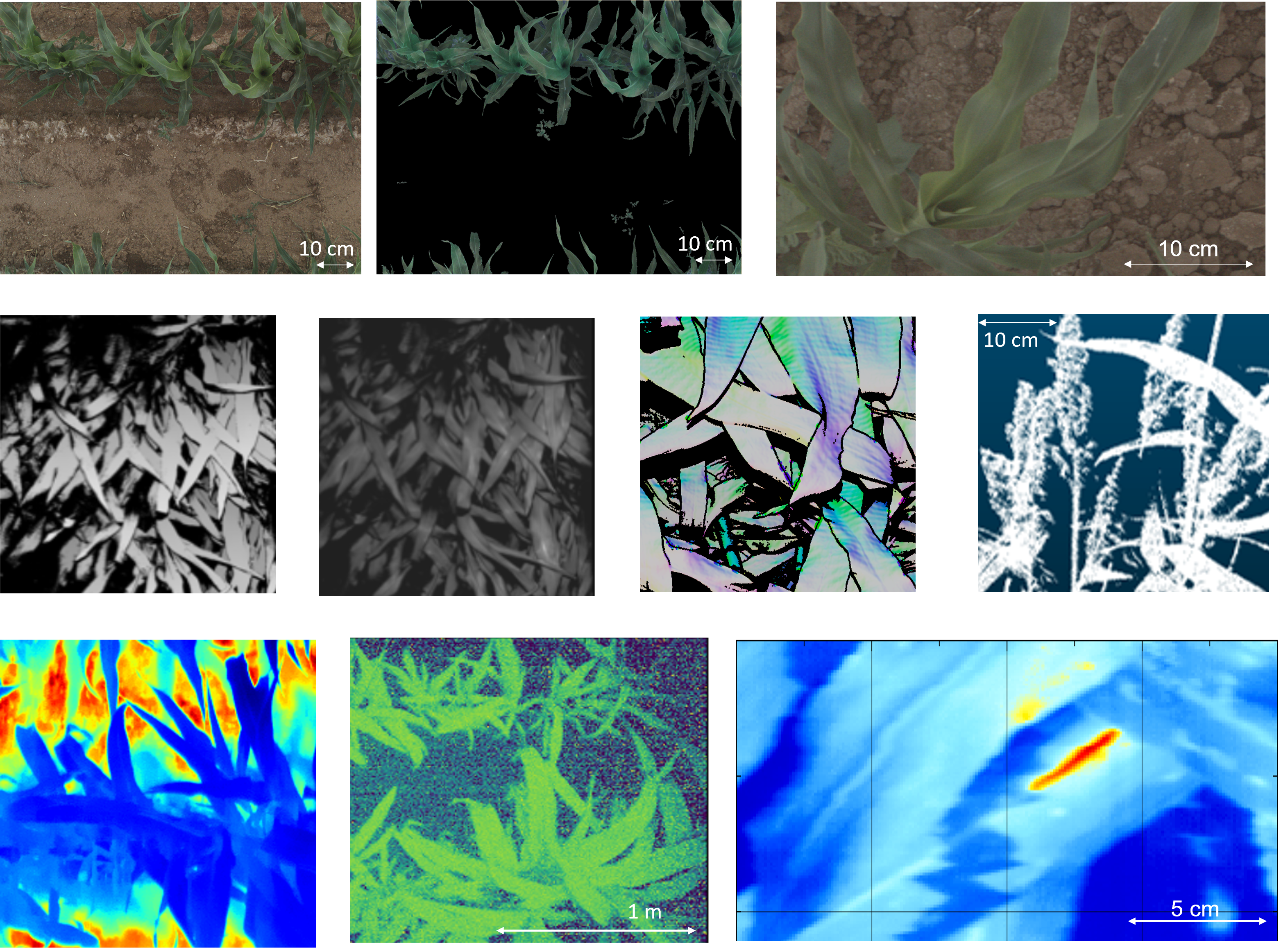}
\caption{\textbf{Example data from the TERRA-REF gantry system.} (top left) RGB data (center top) RGB data with soil masked (top right) close up of RGB data. 3D-scanner data (center row, right to left) depth, reflectance, surface normals, and point cloud data produced by the 3D scanner. (bottom row, right to left) FLIR thermal image with transpiring leaves shown as cooler than soil; F\textsubscript{v}/F\textsubscript{m} derived from active fluorescence PSII sensor providing a measure of photosynthetic efficiency; the reflectance of light at 543nm wavelength measured by the VNIR hyperspectral camera. Because these are two-dimensional representations of three-dimensional systems, all scale bars are approximate.}
\label{fig:example_sensor_data}
\end{figure*}

The TERRA-REF field scanner platform utilizes a sensor suite of co-located instruments (Figure~\ref{fig:sensors} and Table~\ref{tab:sensors}).
The TERRA-REF reference dataset includes several data types (Figures ~\ref{fig:example_sensor_data} and ~\ref{fig:available_sensor_data}, Table~\ref{tab:sensor-data-products}) including raw and processed outputs from sensors, 
environmental sensor measurements, manually measured and computationally derived phenotypes, and raw and processed genomics datasets \cite{LeBauer2020-op}.
Extensive contextual measurements and metadata include sensor information and extensive documentation for each of the sensors, the field scanner, calibration targets, and the results of sensor validation tests \cite{LeBauer2020-op}.

In addition to raw sensor data, the first release of TERRA-REF data includes derived sensor data products in enhanced formats including calibrated and georeferenced images and point clouds (Table ~\ref{tab:sensor-data-products}). 
Many of the data products are provided in formats that follow Open Geospatial Consortium (OGC) standards and work with GIS software.

\subsection{Sensors}
\label{sec:sensors}

Sensors available on the TERRA-REF field scanner include snapshot and line-scan imaging, multi-spectral radiometric, and environmental sensors. 
Table ~\ref{tab:sensors} and Figure ~\ref{fig:sensors}) provide a high level overview of the sensors deployed on this system. 
Full documentation and metadata for each sensor as well as the configuration and geometry of the sensor box are provided as metadata alongside the TERRA-REF data release.

\begin{table*}[hbt]
\caption{Summary of TERRA-REF sensor instruments.}
\label{tab:sensors}
\resizebox{\textwidth}{!}{\begin{tabular}{@{}lll@{}}
\toprule
Sensor Name                             & Model                                  & Technical Specifications                                                                          \\ \midrule
\textbf{Imaging Sensors}                         &                                        &                                                                                                   \\
Stereo RGB Camera                       & Allied Vision Prosilica GT3300C        &                                                                                                   \\
Laser Scanner                           & Custom Fraunhofer 3D                & Spatial Resolution: 0.3 to 0.9 mm                                                                 \\
Thermal Infrared                        & FLIR A615                         &Thermal Sensitivity: $\leq$50mK @ 30 $^\circ$C \\
PS II Camera                            & LemnaTec PS II Fluorescence Prototype & Illumination 635nm x 4000 $\mu$mol/m$^2$/s, Camera 50 fps\\
\textbf{Multi-spectral Radiometers}              &                                        &                                                                                                   \\
Dedicated NDVI Multispectral Radiometer & Skye Instruments SKR 1860D/A           & 650 nm, 800 nm $\pm$ 5 nm; 1 down, 1 up                                                             \\
Dedicated PRI Multispectral Radiometer  & Skye Instruments SKR 1860ND/A          & 531nm +/- 3nm; PRI = Photochemical Reflectance Index                                              \\
Active Reflectance                      & Holland Scientific Crop Circle ACS-430 & 670 nm, 730 nm, 780 nm                                                                            \\
\textbf{Hyper-spectral Cameras}                  &                                        &                                                                                                   \\
VNIR Hyperspectral Imager               & Headwall Inspector VNIR                & 380-1000 nm @ 2/3 nm resolution                                                                   \\
SWIR Hyperspectral Imager               & Headwall Inspector SWIR                & 900-2500 nm @ 12 nm resolution                                                                   \\
\textbf{Environmental Sensors }                  &                                        &                                                                                                   \\
Climate Sensors                         & Thies Clima 4.9200.00.000              &                                                                                                   \\
VNIR Spectroradiometer                  & Ocean Optics STS-Vis                   & Range: 337-824 nm @ 1/2 nm                                                                        \\
VNIR+SWIR Spectroradiometer             & Spectral Evolution PSR+3500            & Range 800-2500nm @3-8 nm; Installed 2018                                                          \\
PAR Sensor                              & Quantum SQ–300                         & Spectral Range 410 to 655 nm                                                                      \\ \bottomrule
\end{tabular}}
\end{table*}

\begin{table*}[hbt]
\centering
\caption{Summary of the sensor data products included in the first release of TERRA-REF data.}
\label{tab:sensor-data-products}
\resizebox{0.7\textwidth}{!}{\begin{tabular}{@{}llllll@{}}
\toprule
\textbf{Data Product} & \textbf{Sensor}     & \textbf{Algorithm} & \textbf{File Format} & \textbf{Plot Clip} & \textbf{Full Field} \\ \midrule
Environment           & Thies Clima         & envlog2netcdf      & netcdf               & NA                 & NA                  \\
Thermal Image         & FLIR                & ir\_geotiff        & geotiff              & +                  &                     \\
Point Cloud           & Fraunhofer Laser 3D & laser3d\_las       & las                  & +                  &                     \\
Point Cloud           & Fraunhofer Laser 3D & scanner3DTop       & ply                  &                    &                     \\
Images Time-Series    & PSII Camera         & ps2png             & png                  &                    &                     \\
Color Images          & RGB Stereo          & bin2tiff           & geotiff              & +                  & +                   \\
Plant Mask            & RGB Stereo          & rgb\_mask          & geotiff              &                    & x                   \\ \bottomrule
\end{tabular}}
\end{table*}

\section{Computer Vision and Machine Learning Problems}

There are a variety of questions that the TERRA-REF dataset could be used to answer that are of high importance to the agricultural and plant science communities, while also posing extremely interesting and challenging computer vision and machine learning problems. In this section, we consider example research areas or topics within computer vision and discuss the relevant agricultural and plant science questions those research communities could help address using the TERRA-REF data.

\paragraph{\textbf{Measurement, Prediction and Causal Inference.}} The TERRA-REF sensor data can be used to drive development of vision-based algorithms for fundamental problems in plant phenotyping, such as making measurements of plant height, leaf length, flower counting, or estimating environmental stress. Additional challenges include attempting to predict \textit{end of season} phenotypes, such as end of season yield, from early season sensor data -- an accurate predictor of end of season yield from early season visual data, for example, could help growers and breeders invest resources only in the most promising of candidate crops. There are additional opportunities to investigate the causal relationship between genotypes or environmental conditions and their expressed phenotypes, as the TERRA-REF dataset includes both comprehensive genetic information, as well as high temporal resolution environmental information. The TERRA-REF data contain over sixty hand measurements that could be used to train models from one or more sensors. In addition, there are opportunities to train models that predict plot-level phenotypes measured by an expensive sensor with a less expensive sensor. Further, many events including insect damage, heat stress, and plant lodging (falling) could be labeled in new images.    

\paragraph{\textbf{Fine Grained Visual Categorization.}} The TERRA-REF data is a rich source of visual sensor data collected from crop species that are visually similar. Differentiating between data with low inter-class variance is an interesting categorization challenge, requiring visual models that learn the fine-grained differences between varieties of the same crop.

\paragraph{\textbf{Transfer Learning.}} There are a variety of interesting transfer learning challenges of utmost importance to the agricultural and plant science communities, including discovering approaches that generalize across sensors, across crops, or across environmental conditions. The TERRA-REF data additionally presents an opportunity to help solve the greenhouse-to-field gap, where models that perform well in greenhouse conditions tend to not generalize to field conditions; because the TERRA-REF data includes both greenhouse and field data for the exact same varieties, researchers in transfer learning could help build models that bridge this gap.

\paragraph{\textbf{Multi-sensor Integration.}} The TERRA-REF data includes data captured from a variety of visual sensors (described in Section~\ref{sec:sensors}). These sensors have similar, but not identical, viewpoints from within the gantry box, may not have captured data at the exact same time, and may have captured different perspectives of the same part of the field on different days. This presents interesting challenges in terms of how to incorporate information across the various sensors, and how to work with time-series data that is not necessarily well-aligned or continuously captured.

\paragraph{\textbf{Explainable Models.}} All too often in machine learning research, datasets and models are built solely to drive the development of machine learning algorithms. When building models to answer questions like ``should I cut this plant down because it won't produce sufficient yield?'' or ``is this plant under environmental stress?,'' it is important not just to have maximally accurate models but to also understand \textit{why} the models make the determinations that they make. This makes the TERRA-REF data, and the biologically relevant questions it supports, an excellent opportunity to drive development of new approaches for explainable machine learning, conveying the decisions made by algorithms to non-machine learning experts.

\paragraph{\textbf{Information Content.}} The TERRA-REF field scanner and sensors represent a substantial investment, and it is still not clear which sensors, sensor configurations, and spatial and temporal resolutions are useful to answer a particular question. Presently, much less expensive sensors and sensing platforms are available \cite{Casto2021-fs,Atkinson2018-lc}. What do we gain from the 1mm spatial resolution on this platform relative to unoccupied aerial systems (UAS) that are quickly approaching 1cm spatial resolution? Or, which subset of hyperspectral wavelengths provide the most useful information? Can we predict the useful parts of a hyperspectral image from RGB images? Or get most of the information from a multispectral camera with a half-dozen bands? At the outset, the team recognized that this configuration would be oversampling the plant subjects, but it wasn't clear what the appropriate resolutions or most useful sensors would be. 

\paragraph{\textbf{Overall Challenges.}} Within all of these topic areas in computer vision and machine learning, the challenges that must be addressed require addressing interesting questions such as determining the most appropriate sensors and data processing choices for specific questions, addressing difficult domain transfer issues, considering how to integrate noisy side channels of information, such as genetic information that may conflict or conflate with each other, or dealing with nuisance parameters like environmental or weather variations that simultaneously influence plant subjects and the sensor data content.

\section{Algorithm Development.}

The process of converting raw sensor outputs into usable data products required geometric, radiometric, and geospatial calibration. In this regard, each sensor presented its own challenges. Combining these steps into an automated computing pipeline also represented a substantial effort that is described by Burnette \etal \cite{Burnette2018-ag}. 

Radiometric calibration was particularly challenging, owing that many images contain both sunlit and shaded areas. 
In the case of hyperspectral images, the white sensor box and scans spread out over multiple days confounded an already challenging problem. 
Radiometric calibration of images taken by the two hyperspectral cameras exemplifies these challenges, and a robust solution is described by Sagan \etal
\cite{sagan2021data} and implemented in \cite{jeromemao_2019_3406312}.
Even processing images from an RGB camera was challenging due to fixed settings resulting in high variability in quality and exposure, requiring the novel approach described by Li \etal \cite{Li_2019_CVPR_Workshops}.
Herritt \etal \cite{herritt2020chlorophyll,herritt2021flip} demonstrate and provide software used in analysis of a sequence of images that capture plant fluorescence response to a pulse of light.

Most of the algorithms used to generate data products have not been published as papers but are made available on GitHub (\url{https://github.com/terraref}); code
used to release the data publication in 2020 is available on Zenodo \cite{craig_willis_2020_3635853, david_lebauer_2020_3635863, max_burnette_2020_3635849, max_burnette_2019_3406304, max_burnette_2019_3406311, jeromemao_2019_3406312, max_burnette_2019_3406318, max_burnette_2019_3406332, max_burnette_2019_3406329, max_burnette_2019_3406335, david_lebauer_2020_3661373}. 

Pipeline development continues to support ongoing use of the field scanner as well as more general applications in plant sensing pipelines. 
Recent advances have improved pipeline scalability and modularity by adopting workflow tools and making use of heterogeneous computing environments.
The TERRA-REF computing pipeline has been adapted and extended for continuing use with the Field Scanner with the new name "PhytoOracle" and is available at \url{https://github.com/LyonsLab/PhytoOracle}. Related work generalizing the pipeline for other phenomics applications has been released under the name "AgPipeline" \url{https://github.com/agpipeline} with applications to aerial imaging described by Schnaufer \etal \cite{Schnaufer2020}.  
All of these software are made available with permissive open source licenses on GitHub to enable access and community development.

\begin{figure*}[hbt]
\centering
\includegraphics[width=0.8\textwidth]{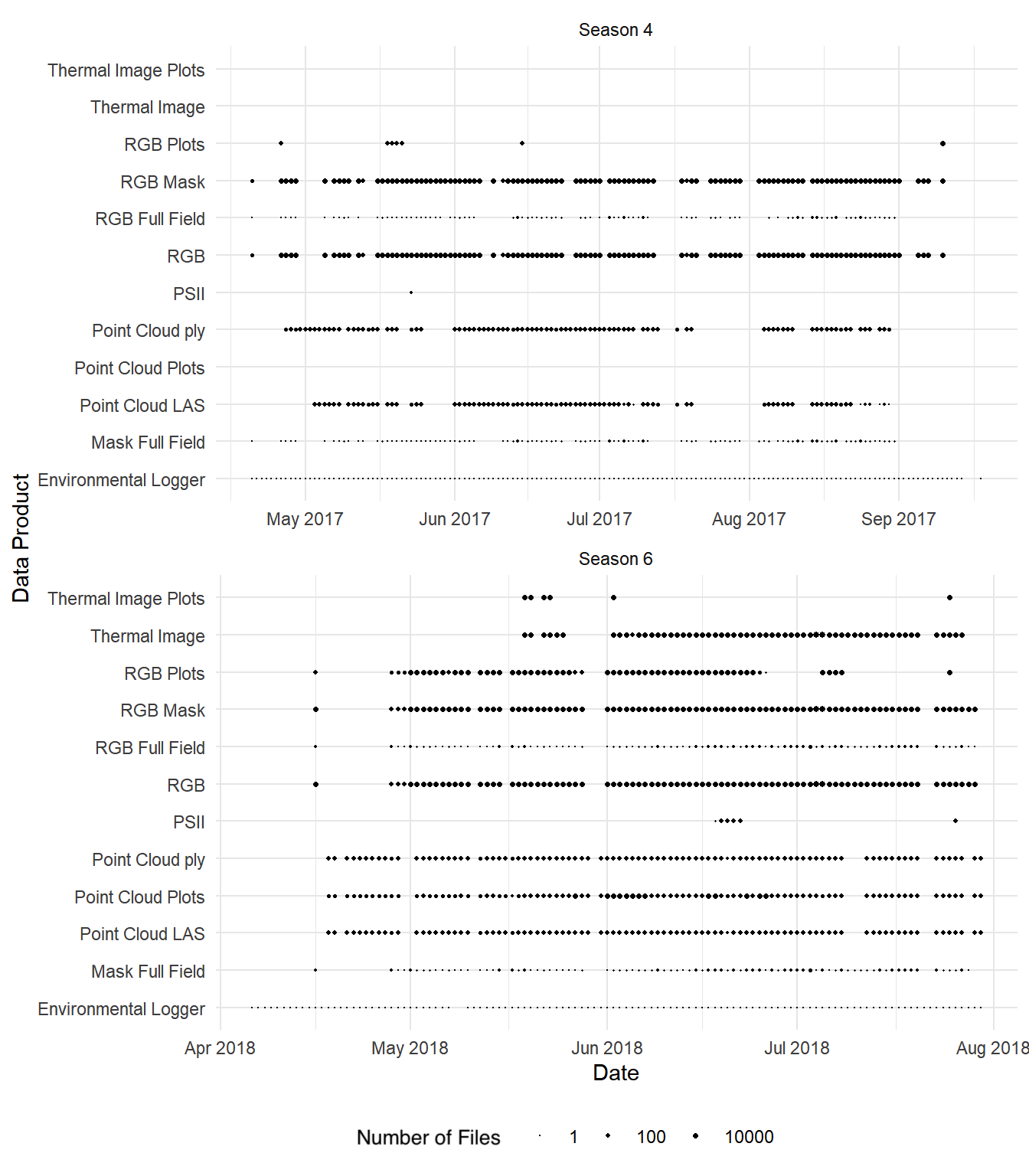}
\caption{Summary of public sensor datasets from Seasons 4 and 6. Each dot represents the dates for which a particular data product is available, and the size of the dot indicates the number of files available.}
\label{fig:available_sensor_data}
\end{figure*}

\section{Uses to date.} TERRA-REF is being used in a variety of ways.
For example, hyperspectral images have been used to measure
soil moisture \cite{babaeian2019new}, but the potential to predict leaf
chemical composition and biophysical traits related to photosynthesis
and water use are particularly promising based on prior work \cite{Serbin2012-gt, Serbin2015-pm}.

\paragraph{\textbf{Plant Science and Computer Vision Research.}} 
A few projects are developing curated datasets for specific machine learning
challenges related to classification, object recognition, and
prediction. 

We currently know of at least three datasets curated for CVPPA 2021. The Sorghum-100 dataset was created to support development of algorithms that can classify sorghum varieties from RGB images from Ren \etal \cite{Ren_2021_CVPR}. Another set of RGB images curated for the
\href{https://www.kaggle.com/c/sorghum-biomass-prediction/overview/iccv-2021-cvppa}{Sorghum
Biomass Prediction Challenge} on Kaggle was developed with the goal of developing methods
to predict end of season biomass from images taken of different sorghum
genotypes over the course of the growing season.
Finally, RGB images from the TERRA-REF field scanner in Maricopa accounted for
250 of the 6000 1024x1024 pixel images in the
\href{https://www.aicrowd.com/challenges/global-wheat-challenge-2021}{Global Wheat Head Dataset 2021}
\cite{david2021global}. The goal of the Global Wheat Challenge 2021
on AIcrowd is to develop an algorithm that can identify
wheat heads from a collection of images from around the world that
represent diverse fields conditions, sensors, settings, varieties, and
growth stages.

Most of the research applications to date have focused on analysis of plot-level phenotypes and genomic data rather than the full resolution sensor data.

\section{Data Access}

\paragraph{\textbf{Public Domain Data.}}

A curated subset of the TERRA-REF data was released to the public domain in 2020 (Figure ~\ref{fig:available_sensor_data}) \cite{LeBauer2020-op}. These data are intended to be re-used and are accessible as a combination of files and databases linked by spatial, temporal, and genomic information. In addition to providing open access data, the entire computational pipeline is open source, and we can assist academic users with access to high-performance computing environments.

The total size of raw (Level 0) data generated by these sensors is 60 TB. Combined, the Level 1 and Level 2 sensor data products are 490 TB. 
This size could be substantially reduced through compression and removal of duplicate data. For example, the same images at the same resolution appear in the georeferenced Level 1 files, the full field mosaics, and the plot-level clip.

\paragraph{\textbf{Other Data Available.}}

The complete TERRA-REF dataset is not publicly available because of the effort and cost of processing, reviewing, curating, describing, and hosting the data.
Instead, we focused on an initial public release and plan to make new datasets available based on need.
Access to unpublished data can be requested from the authors, and as data are curated they will be added to subsequent versions of the public domain release (\url{https://terraref.org/data/access-data}).

In addition to hosting an archival copy of data on Dryad \cite{LeBauer2020-op}, the
documentation includes instructions for browsing and accessing these
data through a variety of online portals. These portals provide access
to web user interfaces as well as databases, APIs, and R and Python
clients. In some cases it will be easier to access data through these
portals using web interfaces and software libraries.

The public domain data is archived on Dryad, with the exception of the large sensor data files. 
The Dryad archive provides a catalog of these files that can be accessed via Globus or directly on the host computer at the National Center for Supercomputing Applications.  

\section{Acknowledgements}
The work presented herein was funded in part by the Advanced Research Projects Agency-Energy (ARPA-E),
U.S. Department of Energy, under Award Numbers DE-AR0000598 and DE-AR0001101, and the National Science Foundation, under
Award Numbers 1835834 and 1835543.

{\small
\bibliographystyle{ieee_fullname}
\bibliography{egbib}
}

\end{document}